\documentclass[conference]{IEEEtran}
\usepackage{times}

\usepackage[numbers]{natbib}
\usepackage{multicol}
\usepackage[bookmarks=true]{hyperref}
\usepackage{algorithm}
\usepackage{amsmath}
\usepackage{xcolor}
\usepackage{float}
\usepackage{amssymb}
\usepackage{booktabs}
\usepackage{graphics} 
\usepackage{graphicx} 
\usepackage[noend]{algpseudocode}
\usepackage[font=small,labelfont=bf]{caption}

\hypersetup{
  colorlinks,
  citecolor= violet,
  linkcolor=blue,
  urlcolor=blue}
\begin{document}

\title{Optimal Transport-Assisted Risk-Sensitive Q-Learning}




%
\author{\authorblockN{Zahra Shahrooei and
Ali Baheri}
\authorblockA{Rochester Institute of Technology\\ Email: {zs9580, akbeme}@rit.edu}
}


\maketitle

\begin{abstract}

The primary goal of reinforcement learning is to develop decision-making policies that prioritize optimal performance without considering risk or safety. In contrast, safe reinforcement learning aims to mitigate or avoid unsafe states. This paper presents a risk-sensitive Q-learning algorithm that leverages optimal transport theory to enhance the agent safety. By integrating optimal transport into the Q-learning framework, our approach seeks to optimize the policy's expected return while minimizing the Wasserstein distance between the policy's stationary distribution and a predefined risk distribution, which encapsulates safety preferences from domain experts. We validate the proposed algorithm in a Gridworld environment. The results indicate that our method significantly reduces the frequency of visits to risky states and achieves faster convergence to a stable policy compared to the traditional Q-learning algorithm. 

\end{abstract}

\IEEEpeerreviewmaketitle

\section{Introduction}

Reinforcement learning (RL) is a general framework for decision-making which enables agents to learn and interact with unknown environments. RL has achieved remarkable success across various domains, including robotics, aviation systems, and healthcare \cite{kober2013reinforcement, razzaghi2022survey}. Traditional RL techniques primarily focus on designing policies that maximize reward in a Markov decision process (MDP) and neglect the potential risks associated with the actions taken. This has led to the development of safe RL algorithms, which aim to maximize performance while guaranteeing or encouraging safety.





In the past few years, researchers have introduced various strategies to ensure safety in RL. These strategies encompass constrained RL techniques \cite{achiam2017constrained}, the implementation of safety layers or shielding mechanisms \cite{alshiekh2018safe,baheri2020deep}, and the use of formal methods \cite{bansal2022specification}. Comprehensive reviews of these safe RL methodologies can be found in \cite{garcia2015comprehensive,liu2021policy}.
Safe RL methods comprise both model-based and model-free approaches \cite{gu2022review}. Model-based methods employ explicit models of the environment to predict future states and assess safety more accurately. \cite{chow2018risk} proposed policy gradient and actor-critic methods based on optimization theory to enhance risk-sensitive RL performance. Their approach incorporates conditional value at risk and chance-constrained optimization to ensure safety. \cite{berkenkamp2017safe} presents a safe model-based RL algorithm by employing Lyapunov functions to guarantee stability with the assumptions of Gaussian process prior; Generally, Lyapunov functions are hand-crafted, and it is challenging to find a principle to construct Lyapunov functions for an agent's safety and performance. Furthermore, some safe RL studies use model predictive control to make robust decisions in constrained MDPs. In these approaches, the system inputs are designed by solving an optimization problem that depends explicitly on the model of the system or use system identification in the case that model is not available \cite{zanon2020safe, tsiamis2019finite}. Shielding is another technique to ensure safe decision making \cite{bloem2015shield, alshiekh2018safe}. By augmenting an RL agent with a
shield, at every time step, unsafe actions are blocked by the
shield and the learning agent can only take safe actions.

On the other hand, model-free safe RL has attracted significant interest from researchers because it can be applied across various domains without requiring knowledge of model dynamics. In model-free methods, safety is typically integrated into the learning process by modifying the reward function to include safety constraints or by using external knowledge to guide exploration. One of the most important safe RL methods, primal-dual methods convert the original constrained
optimization problem to its dual form by introducing a set of Lagrange multipliers \cite{yang2022cup}. Techniques like constrained policy optimization (CPO) and trust region policy optimization (TRPO) with safety constraints are used to ensure that policies remain within safe bounds during training \cite{achiam2017constrained, yang2020projection}. Recently, primal-based methods have become a strong alternative for primal-dual-based approaches. \cite{xu2021crpo} presented constraint-rectified policy optimization (CRPO), the first primal-based framework for safe RL; however, it may have experienced oscillations between reward and safety optimization, which could negatively impact its performance. To address this issue, projection CRPO \cite{gu2024balance} implemented a soft policy optimization technique using gradient manipulation theory which outperforms CRPO. Additionally, There exist studies on safe RL algorithms in model-free settings using Lyapunov functions \cite{chow2019lyapunov}, control barrier functions \cite{hasanbeig2020cautious}, and formal methods \cite{murugesan2019formal}.

Finally, several earlier works have tackled safe exploration problem in an environment that is unknown a priori \cite{sui2015safe, koller2018learning, dalal2018safe, pecka2014safe}. This kind of problem setting is ideal for cases such as a robot exploring an uncertain environment. Some safe exploration techniques take advantage from different kinds of prior knowledge such as initializing search using the prior knowledge \cite{okawa2022safe} or learning from human demonstrations \cite{ramirez2023safe}.

Originating from the work of Gaspard Monge and later advanced by Leonid Kantorovich, Optimal transport (OT) theory has became well-known in various domains due to its ability to measure and optimize the alignment between probability distributions \cite{villani2009optimal}. There have been several applications of OT in RL \cite{baheri2023understanding, zhang2018policy, pacchiano2020learning, queeney2023optimal, zare2023leveraging}. It is more robust than other measures, such as Kullback-Leibler divergence, because it captures both the geometric and probabilistic differences between distributions which makes it particularly effective in scenarios where data distributions differ significantly in shape or support \cite{santambrogio2015optimal}. \cite{queeney2023optimal} applies OT theory to develop a safe RL framework that incorporates robustness through an OT cost uncertainty set. This approach constructs worst-case virtual state transitions using OT perturbations which improves safety in continuous control tasks compared to standard safe RL methods. \cite{nkhumise2024measuring} uses OT to develop an exploration index that quantifies the effort of knowledge transfer during RL as a sequence of supervised learning tasks to compare the exploration behaviors of various RL algorithms. Furthermore, OT has been employed in multi-agent RL to enhance the efficiency, coordination, and adaptability of agents. OT's Wasserstein metric facilitates policy alignment among agents and optimizes distributed resource management \cite{baheri2024synergy}.



In this paper, we build upon our previous paper \cite{baheri2023risk} which presents a risk-sensitive Q-learning algorithm that incorporates optimal transport theory to enhance agent safety during learning. The proposed
Q-learning framework minimizes the Wasserstien distance between the policy's stationary distribution and a predefined risk distribution. Wasserstien distance effectively measures the policy's safety level. A lower distance suggests that the policy aligns well with the safety preferences, meaning the agent tends to visit safer states more frequently and avoids higher-risk ones. 


\section{Preliminaries}

This section provides a review on MDPs, which serve as the foundational framework for modeling decision-making problems, Q-learning algorithm and basic principles of OT theory.

\subsection{Markov Decision Processes}
MDPs formally describe an environment for reinforcement learning where the environment is fully observable. A finite MDP is defined by a tuple $\mathcal{M} = (\mathcal{S}, \mathcal{A}, T, \gamma, \rho)$, where $\mathcal{S} = \{s_1, \ldots, s_{n}\}$ is a finite set of states, $\mathcal{A}$ is a finite set of actions available to the agent, and $T$ is the transition probability function such that $T(s' \mid s, a)$ describes the probability of transitioning from state $s$ to $s'$ given a particular action $a$. $\gamma \in \left ( 0,1 \right )$ is a discounting factor, and $\rho$ specifies the initial probability distribution over the state space $\mathcal{S}$. The agent behavior is defined by a policy that maps states to a probability distribution over the actions $\pi : \mathcal{S} \times A \rightarrow [0, 1]$, and its objective is to maximize the expected discounted return of rewards $G_t = \sum_{k=0}^{\infty} \gamma^k r_{t+k+1}$.

\subsection{Q-learning Algorithm}

Q-learning is a a foundational model-free, value-based, off-policy algorithm that interacts with environment iteratively to approximate the state-action value function $Q(s, a)$. The Q-values are updated based on the Bellman optimality equation \cite{watkins1992q}:

\begin{equation}
Q(s, a) \leftarrow Q(s, a) + \alpha \left( r + \gamma \max_{a'} Q(s', a') - Q(s, a) \right) \label{eq:1}
\end{equation}
where $\alpha \in [0,1]$ is the learning rate, $r$ represents the immediate reward received after transitioning from state $s$ to state $s'$ by taking action $a$. This process continues until the Q-values converge which indicates the optimal policy has been found. To balance exploration and exploitation during learning, Q-learning often employs $\epsilon$-greedy and $\epsilon$-decaying strategies. The $\epsilon$-greedy strategy allows the agent to choose a random action with probability $\epsilon$ and the action that maximizes the Q-value with probability $1 - \epsilon$. This ensures that the agent explores the environment sufficiently while still exploiting known information to maximize rewards. The $\epsilon$-decaying strategy gradually decreases the value of $\epsilon$ over time which reduces the exploration rate as the agent becomes more confident in its learned policy.

\subsection{Optimal Transport Theory}

OT theory provides a mathematical tool for transporting mass between probability distributions efficiently \cite{solomon2018optimal}. In the discrete case, we consider two distributions $P = \sum_{i=1}^n p_i \delta_{s_i}$ and $Q = \sum_{j=1}^n q_j \delta_{s_j}$ over state space $\mathcal{S}$, where $\delta_{s_i}$ and $\delta_{s_j}$ are Dirac delta functions at $s_i$ and $s_j$, respectively.

The cost of moving mass from position $i$ to position $j$ is given by $C_{ij}$. The transport plan is represented by a matrix $T \in \mathbb{R}^{n \times n}$, where $T_{ij}$ indicates the amount of mass transported from $s_i$ to $s_j$. Hence, the total cost of a transport plan is given by:

\begin{equation}
 \sum_{i=1}^n \sum_{j=1}^n T_{ij} C_{ij} \label{eq:2}
\end{equation}
The OT problem involves finding the transport plan $T^\ast$ that minimizes this cost while ensuring the marginal distributions of the transport plan match $P$ and $Q$:

\begin{equation}
\begin{aligned}
\arg \min_{\Gamma \in \mathbb{R}^{n \times n}} \quad & \sum_{i=1}^n \sum_{j=1}^n T_{ij} C_{ij} \\
\text{subject to} \quad & \sum_{j=1}^n T_{ij} = p_i, \quad \forall i \in \{1, \ldots, n\} \\
& \sum_{i=1}^n T_{ij} = q_j, \quad \forall j \in \{1, \ldots, n\} \\
& T_{ij} \geq 0, \quad \forall i, j
\end{aligned}
\label{eq:3}
\end{equation}
where $\Gamma$ is the set of all transport plans between $P$ and $Q$. Letting $T^\ast$ denote the solution to the above optimization problem, the Wasserstein distance between $P$ and $Q$ is defined as:
\begin{equation}
W_{p}(P, Q) = \left(  \langle T^{\ast}, C \rangle \right)^{\frac{1}{p}} \label{eq:4}
\end{equation}

\section{Risk-sensitive Reinforcement Learning with Optimal Transport}

We consider an RL agent operating within an MDP environment.
We also assume that a reference risk distribution $P^{s} : S \to [0,1]$ which represents the safety preferences over the state space $S$ is provided by domain experts. This distribution quantifies the relative safety of each state, with higher probabilities assigned to safer states and lower probabilities to riskier states. The state distribution $P^{\pi }$ appears as the stationary distribution for the Markov chain produced by a given policy $\pi$ in the MDP. This distribution represents the long-term behavior of the agent under current policy $\pi$, specifically describing the proportion of time the agent will spend in each state regardless of the initial state. The goal is to obtain a policy that not only optimizes expected returns but also reduces the Wasserstein distance between the policy's stationary distribution and the reference risk distribution. Therefore, we aim to determine the transport plan $T^{\ast}$ that gives the most cost-efficient way of transforming stationary distribution (source) into risk distribution (target). Once $T^{\ast}$ is determined for any pair of states $(s, s')$, the value of $T^{\ast}$ tells us what proportion of source distribution $P^{\pi }$ at $s$ should be transferred to $ s'$, in order to reconfigure $P^{\pi }$ to $P^{s}$.

In this study, we consider the cost function 
$c : \mathcal{S} \times \mathcal{S} \rightarrow \mathbb{R}$ which quantifies the cost of transporting probability mass from state $s$ to state $s'$, to be the squared Euclidean distance, $c(s, s') = \|s - s'\|^2$. Therefore, the cost for transition from state $s$ to $s'$, is obtained by $C\left ( s,s' \right )= T^{\ast}\left (  s,s'\right )c\left ( s,s' \right )$. Upon determining $C\left ( s,s' \right )$ for any pair of states $(s, s')$, we can modify the update rule for Q-learning framework in Eq. \ref{eq:1} as follows:

\begin{equation}
\begin{split}
Q(s, a) &\leftarrow Q(s, a) + \alpha \Big[ r + \gamma  \max_{a'} Q(s', a') - Q(s, a)\\
&\quad + \beta \cdot C(s, s') \Big] \label{eq:5}
\end{split}
\end{equation}
where $\beta$ is a sensitivity parameter that balances the trade-off between the expected rewards and the cost of transitioning from state $s$ to $s'$. The incorporation of the OT term into the Q-learning update rule influences the learning algorithm to prefer transitions that are safer according to the transport plan. 

\begin{algorithm}[!b]
\caption{Optimal Transport-Assisted Q-Learning}\label{alg:qot}
\begin{algorithmic}[1]

\Require  sensitivity parameter $\beta$, risk distribution $P^{s}$,
\State Initialize $Q(s, a) \leftarrow 0$ for all $(s, a)$
\For {$1,\cdots ,n$}:
  \State Initialize state $s_0$
  \While {not terminal}
  \State \parbox[t]{\dimexpr\linewidth-\algorithmicindent}{With probability $\epsilon$ select a random action $a$ \\ 
   Otherwise select $a = \arg\max_{a'} Q(s, a')$}
  \State Take action $a$, observe reward $r$ and next state $s'$
  \State Update $Q(s, a)$ according to Eq. \ref{eq:5}
  \State $C(s, s') \leftarrow 0$
  \State $s \leftarrow s'$
  \EndWhile
  \State Obtain $P^{\pi }$ for updated policy $\pi$
  \State Recompute OT plan $T^\ast$ based on Eq. \ref{eq:3}
  \State Decay $\epsilon$
\EndFor
\end{algorithmic}
\end{algorithm}

\begin{figure}[ht!]
     \centering
     \includegraphics[scale=1.2]{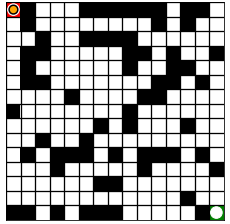}
     \caption{Gridworld environment}\label{Fig 1. }
\end{figure}

The process of OT integration to Q-learning is described in Algorithm \ref{alg:qot}. The algorithm starts by initializing the Q-values with zero values for all state-action pairs. For each episode, the agent interacts with the environment iteratively by selecting actions that maximize the current Q-values perturbed by an exploration factor $\epsilon$, which decays over time to balance exploration and exploitation. After each action, the corresponding Q-value is updated considering both the immediate reward and the corresponding cost which reflects the safety level associated with the transitions. The cost $C\left ( s,s' \right )$ is then updated to zero to ensure that if the agent repeats the same transition, the OT term is not applied multiple times. At the end of each episode, the stationary distribution is recalculated based on the updated policy, and a new optimal transport plan is determined. This process is repeated for a predefined number of episodes or until convergence.

\section{Simulations and Results}

To illustrate the efficiency of our proposed algorithm, we evaluate it on a $15\times 15$ Gridworld with obstacles shown in Figure \ref{Fig 1. }. The agent can perform four
different actions: up, down, right, left. The reward for this environment, for hitting the obstacles is $-10$, for reaching the goal state is $10$, and $-1$ for general movement in the environment. We define the safety distribution of states (except for goal state that has safety probability $1$) to get $-0.3$ for being adjacent to boarders or obstacles. We consider the risk distribution over the states, which reflects the safety preferences to be $1$ for all states. This safety value is decreased by $0.3$ for each adjacency to borders or obstacles. The goal state gets a safety value of $1$. We compare our method with standard Q-learning in $5$ random seeds across $500$ episodes.


\begin{figure}[ht!]
     \centering
     \includegraphics[scale=0.6]{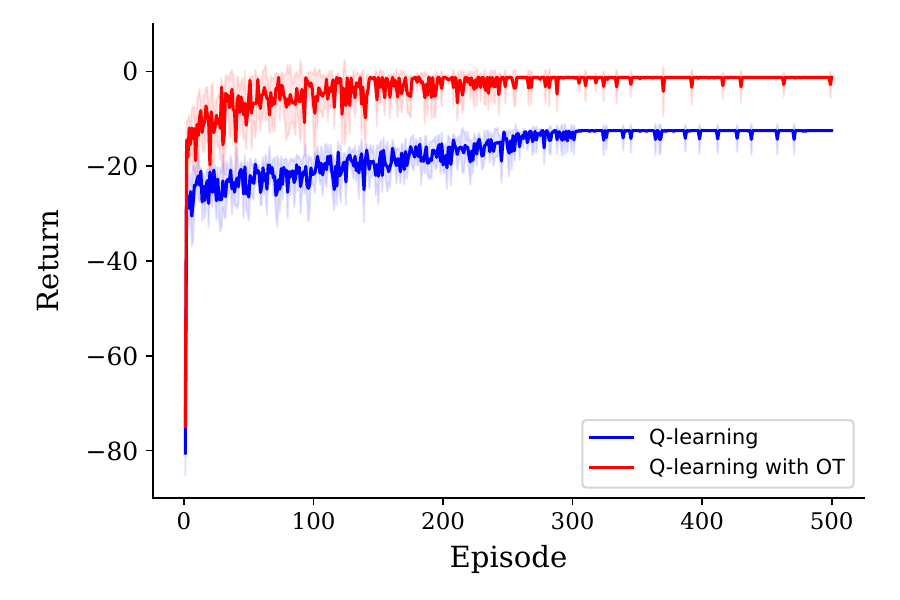}
     \caption{Average return values across $500$ episodes for $5$ random seeds.}\label{Fig 2. }
\end{figure}


Figure \ref{Fig 2. } shows the return values for our algorithm and standard Q-learning over $500$ episodes. As expected, the return for our algorithm is higher than that of standard Q-learning, since we reshaped the rewards with the OT term, which is always a non-negative term. Additionally, our algorithm converges about $150$ episodes faster than traditional Q-learning. This significant improvement in convergence becomes evident from the second episode onwards, when the OT term starts influencing the learning process.

\begin{figure}[ht!]
     \centering
     \includegraphics[scale=0.5]{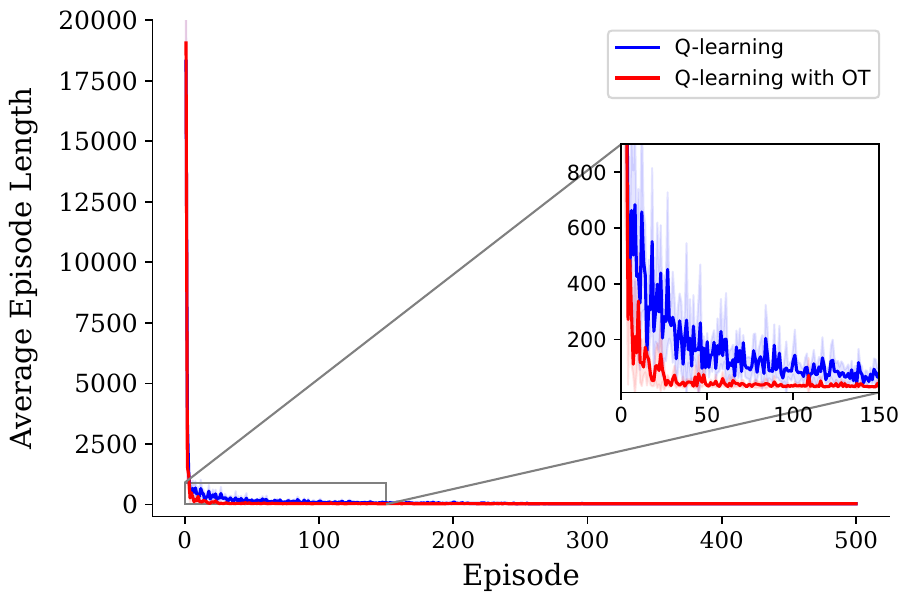}
     \caption{\textcolor{black}{Average episode length for $5$ random seeds.}}\label{Fig 3. } 
\end{figure}

Figure \ref{Fig 3. } illustrates the average length of episodes across the same $500$ episodes. Our algorithm demonstrates a rapid decrease in episode length, indicating that the agent learns to reach the goal state more efficiently over time. This reduction in episode length suggests that the policy is effectively aligning with the safety preference distribution, enabling the agent to avoid obstacles more frequently.

\begin{figure}[ht!]
     \centering
     \includegraphics[scale=0.5]{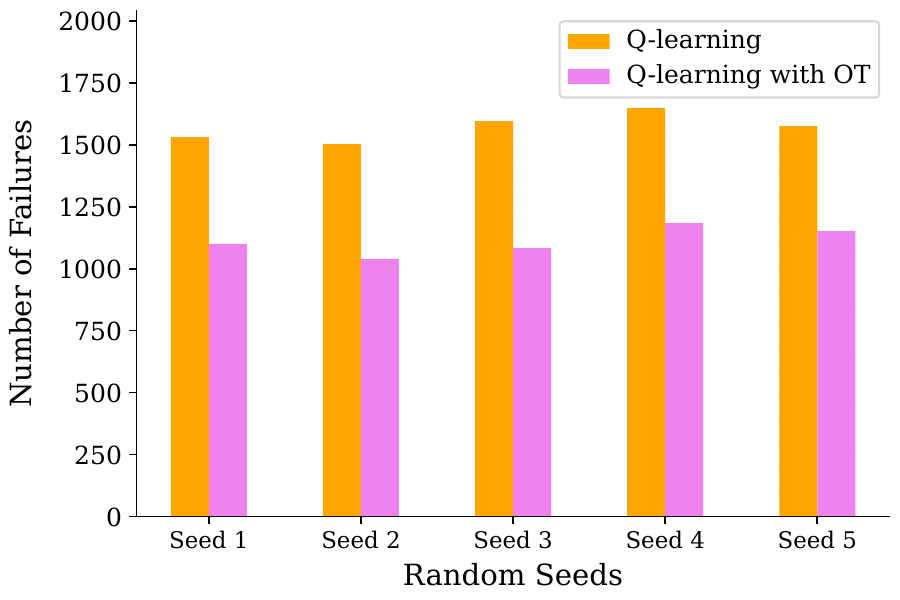}
     \caption{Number of obstacle collisions over $500$ episodes for $5$ random seeds.}\label{Fig 4. }
\end{figure}

In Figure \ref{Fig 4. }, number of collisions with obstacles is depicted. The risk-sensitive Q-learning algorithm exhibits a significant decrease in the number of obstacle collisions compared to the traditional Q-learning algorithm, with the average number of collisions being $30\%$ less across five random seeds. This further confirms the effectiveness of incorporating OT theory into the Q-learning framework, as the agent is directed away from high-risk areas.

\section{Conclusions}\label{sec: Conclusion}

We presented a risk-sensitive Q-learning algorithm that uses optimal transport theory to guide the agent towards safe states by aligning the stationary distribution of the policy with safety preference distribution. Our approach stands out for its ability to considerably cut down on the number of visiting risky states and convergence time. The proposed Q-learning approach, while promising, does have limitations. A primary concern is the dependence on an accurate risk distribution, which can be difficult to define and obtain in complex environments. Additionally, the computational demands of recalculating the optimal transport plan at the end of each episode are significant, which potentially restricts the scalability of this approach for larger state spaces. Future research will focus on addressing these challenges. To mitigate the dependency on predefined safety preferences, we plan to explore methods for learning the safety preference distribution dynamically during training. This could involve using online learning techniques or incorporating feedback from human experts. To reduce computational overhead, we will investigate more efficient algorithms for optimal transport computation and explore approximation methods.



\bibliographystyle{plainnat}
\bibliography{main}

\end{document}